\pdfoutput=1

\documentclass[11pt]{article}
\usepackage{acl}

\usepackage{times}
\usepackage{latexsym}

\usepackage[T1]{fontenc}

\usepackage[utf8]{inputenc}

\usepackage{microtype}

\usepackage{inconsolata}

\usepackage{graphicx}

\DeclareMathAlphabet{\mathbcal}{OMS}{cmsy}{b}{n}
\usepackage{amsmath}
\usepackage{mathrsfs}
\usepackage{comment}

\usepackage{bm}
\usepackage{bbm}
\usepackage{amssymb}
\usepackage{enumitem}
\usepackage{float}
\usepackage{booktabs}

\def\bsq#1{%
\lq{#1}\rq}

\usepackage{subfigure}
\usepackage{graphbox}
\usepackage{svg}
\usepackage{nicefrac}
\usepackage{amsmath}
\usepackage{siunitx}

\usepackage{float}

\newcommand{\vy}{{\bf y}}

\usepackage[T1]{fontenc}
\usepackage[utf8]{inputenc}
\usepackage{microtype}
\usepackage{inconsolata}

\usepackage{graphicx}

\title{A Framework for Fine-Tuning LLMs using Heterogeneous Feedback}

\author{
 \textbf{Ryan Aponte \textsuperscript{1}},
 \textbf{Ryan A. Rossi \textsuperscript{2}},
 \textbf{Shunan Guo\textsuperscript{2}},
 \textbf{Franck Dernoncourt\textsuperscript{2}},
\\
 \textbf{Tong Yu\textsuperscript{2}},
 \textbf{Xiang Chen\textsuperscript{2}},
 \textbf{Subrata Mitra\textsuperscript{2}},
 \textbf{Nedim Lipka \textsuperscript{2}}
\\
\\
 \textsuperscript{1}Carnegie Mellon University,
 \textsuperscript{2}Adobe Research
}

\begin{document}
\maketitle

\begin{abstract}
Large language models (LLMs) have been applied to a wide range of tasks, including text summarization, web navigation, and chatbots. They have benefitted from supervised fine-tuning (SFT) and reinforcement learning from human feedback (RLHF) following an unsupervised pretraining. These datasets can be difficult to collect, limited in scope, and vary in sample quality. Additionally, datasets can vary extensively in supervision format, from numerical to binary as well as multi-dimensional with many different values. We present a framework for fine-tuning LLMs using heterogeneous feedback, which has two main components. First, we combine the heterogeneous feedback data into a single supervision format, compatible with methods like SFT and RLHF. Next, given this unified feedback dataset, we extract a high-quality and diverse subset to obtain performance increases potentially exceeding the full dataset. We conduct extensive experiments to understand the effectiveness of these techniques for incorporating heterogeneous feedback, and demonstrate improvements from using a high-quality and diverse subset of the data. We find that our framework is able to improve models in multiple areas simultaneously, such as in instruction following and bias reduction.
\end{abstract}

\section{Introduction}

LLMs are fine-tuned for a variety of purposes, such as for instruction following in InstructGPT~\cite{ouyang2022training}. The fine-tuning process generally begins with collecting examples of desired model behavior and performing supervised learning. Some models stop at SFT~\cite{vicuna2023}, while InstructGPT follows this by training a reward model based on binary human preference data. The fine-tuned model is then further refined using RLHF, using a signal from the reward model. In the example of InstructGPT, the algorithm used is Proximal Policy Optimization (PPO)~\cite{schulman2017proximal}. For each of these steps, the fine-tuning dataset uses a single form of supervision. 
Fine-tuning datasets exist for a variety of purposes, from training chat-based assistants in OASST~\cite{köpf2023openassistant}, coreference resolution in  WinoGrande~\cite{sakaguchi2019winogrande}, helpfulness, honesty, and harmlessness in Anthropic HHH~\cite{DBLP:journals/corr/abs-2112-09332}, and logical reasoning in OpenPlatypus~\cite{lee2024platypus}. Supervision format varies, from binary preference in Anthropic HHH, to several numerical labels OASST, to a string response in OpenPlatypus. Although fine-tuning has been successful in mitigating the limitations of pretrained LLMs, these methods require data of a single supervision type, restricting the scope of preference data. 
Recent work has filtered fine-tuning datasets to reduce cost and increase quality~\cite{wang2024surveydataselectionllm}. ~\cite{wu2023selfevolveddiversedatasampling} use LLMs to generate embeddings for fine-tuning data which is clustered with k-center-greedy~\cite{sener2018activelearningconvolutionalneural} using an iterative process. ~\cite{kung2023activeinstructiontuningimproving} randomly delete words in prompts and measure how the response probability changes as a measure of the model's uncertainty. ~\cite{li2024quantityqualityboostingllm} outperform Alpaca as evaluated by LLM preference using only 5\% of its fine-tuning data.

We present a framework to use multiple fine-tuning data types, permitting the use of more fine-tuning datasets and fine-tuning for multiple tasks simultaneously. Using multiple datasets enables fine-tuning for different goals simultaneously, such as for logical reasoning and to reduce bias, and provides a more accurate view of human preference by broadening the scope of fine-tuning data. Our framework selects a high-quality and diverse subset of the data to make fine-tuning more effective.

\begin{figure*}
    \centering
    \includegraphics[width=\textwidth]{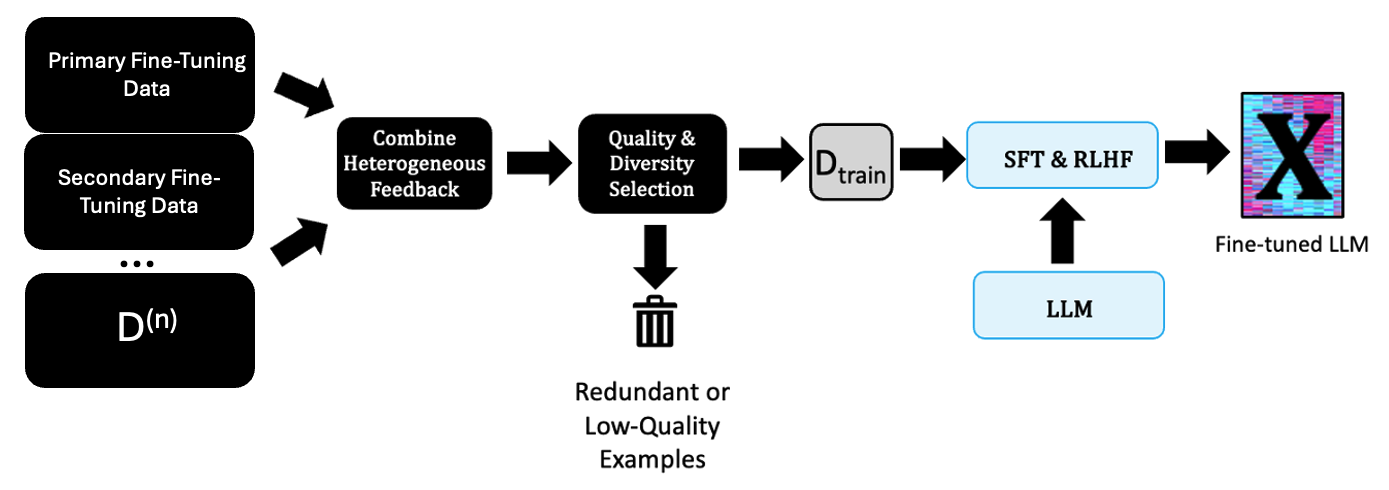}
    \caption{Framework. First, we concatenate the datasets into a dataset of heterogeneous feedback. We then score samples based on quality and prompt diversity, remove a fraction of the samples (a hyperparameter), forming the homogeneous dataset $D_{train}$. Standard fine-tuning methods are then applied to a pre-trained LLM.
    }
    \label{fig:schematic}
\end{figure*}

\section{Framework} \label{sec:framework}

The primary contribution of our framework is to be able to use fine-tuning data of heterogeneous supervision. Figure~\ref{fig:schematic} includes a high-level overview. Our framework utilizes the simplest supervision, such as binary preference, and projects all remaining datasets into that format. Because some data may be redundant in the unified dataset, we filter for quality and diversity to generate $D_{train}$. For simplicity, we use this dataset for both the SFT and RLHF steps of fine-tuning, however this is not a requirement. This generates an LLM fine-tuned with high-quality and diverse data, LLaMA-HD.

\subsection{Primary fine-tuning dataset}\label{sec:primary-finetune}
Given a dataset $\mathcal{D}$ of prompts with two responses using binary preference,
\begin{equation}
    \mathcal{D} = \{ (P_i, A_{i,0}, A_{i,1}\}_{i=1}^M
\end{equation}
where $P$ is the prompt, $A_{i,0}$ and $A_{i,1}$ are answers to the prompt, with $A_{i,0}$ defined as the preferred response to the prompt.
This type of dataset takes the form of binary preference due to two example responses to a single prompt. Examples here do not convey a sense of quality, thus prohibiting ranking.

\subsection{Secondary fine-tuning dataset}\label{sec:secondary-finetune}

Given a dataset $\mathcal{D^*}$ of user-specific prompts and responses (question-answer tuples): 
\begin{equation}
    \mathcal{D^*} = \{(P_i, A_i, \vy_i)\}_{i=1}^N
\end{equation}
where $P_i$ and $A_i$ are the $i$th prompt and response pair, 
respectively, and $\vy_i \in \mathbb{R}^{k}$ is the real-valued vector denoting the score of various labels for that pair. For a dataset of this type to be compatible with our method, it is necessary that there are multiple responses to the same prompt. For example,
\begin{equation}
    (P_i, A_{i'}, \vy_{i'}) \in D^*
\end{equation}
can be the second response to the prompt. This process can be repeated for arbitrarily many datasets. A general method for one axis of supervision is included in Appendix~\ref{sec:appendix-converting}.

\subsection{Simple Unionization for Feedback}
We take $\mathcal{D^*}$ and create a dictionary with prompt as key and responses as a list of all responses to that prompt. This requires at least two responses for each prompt to be considered. We can conduct quality and diversity filtering on these prompts, and then select the preferred response pairs. Once we have tuples containing a prompt, preferred, and non-preferred response, our data from $D^*$ are now in the same format as $D$, so we take the union.

\subsection{Quality Selection}

We infer example quality based on the numerical labels of responses. For datasets with multiple numerical labels, selection of the label is a hyperparameter likely motivated by the purpose of fine-tuning. For example, our experiment uses toxicity as this is related to our objective of reducing bias. For prompts with more than two responses, the highest quality pair of responses are those that vary most in the numerical label. Intuitively, these should give a strong signal to a reward model because one response is strongly preferred. Finally, we can rank and select prompts by the preference difference of their responses.

\subsection{Diversity Selection}
We select for prompt diversity by generating embeddings for each prompt, followed by clustering. Prompts with similar meaning can be considered redundant. We follow OpenPlatypus in using a sentence transformer to generate semantic embeddings to filter datasets~\cite{lee2024platypus}. Embeddings are then clustered using unsupervised methods like k-means. We select the top fraction of responses from each cluster. Both the number of clusters and fraction of each cluster are hyperparameters.

\subsection{Training}
We use the training pipeline from StackLLaMA~\cite{beeching2023stackllama}, which uses LLaMA-7B~\cite{touvron2023llama}. First, we perform SFT on the base model. We then train a reward model using the fine-tuned model. This is followed by RLHF on the fine-tuned model using PPO. Low-Rank Adaptation is used to reduce memory usage and increase parallelization~\cite{hu2021lora}. We select varying fractions of the training dataset, as well as omit filtering, to measure its influence.

\begin{table*}[h]
  \centering
\caption{Quantitative Results. Bolded entries denote highest performance. -S indicates model was fine-tuned with SFT only, -R is SFT followed by RLHF. No filtering was performed on fine-tuning datasets for LLaMA-S and LLaMA-R.}
\label{tab:results}

\begin{tabular}{lccccc}
    \hline
    \textbf{Model} & \textbf{Bias $\downarrow$} & \textbf{Bias (Entropy) $\downarrow$} & \textbf{Bias (Cluster) $\downarrow$} & \textbf{Accuracy $\uparrow$} & \textbf{Similarity $\uparrow$} \\
    \hline
     LLaMA-Base & 0.4585 & \textbf{0.0010} & 3.0393 & 0.9482 & 0.9482 \\
     LLaMA-S & 1.1721 & 0.1553 & 10.3180 & 0.5953 & 0.6553 \\
     LLaMA-R & 0.9247 & 0.0098 & 4.2139 & 0.9457 & 0.9457 \\
     \hline
     LLaMA-HD-0.2-S & 0.4436 & 0.0580 & 4.5856 & 0.9204 & 0.9204 \\
     LLaMA-HD-0.4-S & 0.7798 & 0.0548 & 6.3741 & 0.8788 & 0.9394 \\
     LLaMA-HD-0.6-S & 0.7947 & 0.0407 & 7.5564 & 0.8327 & 0.8927 \\
     LLaMA-HD-1.0-S & 0.4117 & 0.0333 & 3.8851 & 0.9533 & 0.9533 \\
     LLaMA-HD-0.2-R & 0.4330 & 0.0580 & 3.0892 & 0.9482 & 0.9482 \\
     LLaMA-HD-0.4-R & 0.4287 & 0.0548 & 2.9852 & 0.9602 & 0.9508 \\
     LLaMA-HD-0.6-R & 0.4727 & \textbf{0.0010} & \textbf{2.9472} & \textbf{0.9646} & 0.9571 \\
     LLaMA-HD-1.0-R & \textbf{0.3629} & 0.0068 & 3.1570 & 0.9583 & \textbf{0.9583} \\
     \hline
\end{tabular}

\end{table*}

\section{Experimental Setup}\label{sec:experiment-setup}

\subsection{Heterogeneous Datasets}\label{sec:hetero-datasets}
We use three datasets for our experiments: WinoGrande~\cite{sakaguchi2019winogrande} (our primary dataset), OpenAssistant OASST~\cite{köpf2023openassistant} (our secondary dataset), and WinoGender~\cite{zhao2018gender} for testing the generalization of our method. WinoGrande is a coreference resolution dataset developed as a more challenging alternative to the Winograd Schema Challenge~\cite{levesque2012winograd}, as machine learning models exceeded 90\% accuracy on the dataset. WinoGrande has been found to transfer to other wino-style schema challenges, including WinoGender. OASST is a conversation dataset consisting of over 10,000 conversation trees. This dataset has numerical supervision, providing an inherent measure of quality. WinoGender is a dataset testing gender bias in coreference resolution that involve a pair of sentences, one conforming to American gender biases and one against them. Differences in response indicate gender bias.

We fine-tune using either WinoGrande alone with LLaMA-SFT and LLaMA-RLHF, or with a combination of WinoGrande and OASST using our framework. We fine-tune using several subsets of the data, in addition to the dataset without filtering.  For SFT and training the reward model, we treat the pro-bias examples of WinoGender as negative and the anti-bias examples as positive. This follows from the intuition that language models learn human biases, so reductions in bias can be achieved by training models in the opposite direction.

\subsection{Dataset Filtering}\label{sec:exp-selection}

We use the numerical score toxicity in OASST to measure prompt quality. Prompts are ranked based on the difference in toxicity between responses. By reducing toxicity, we may also be able to reduce gender bias. For prompts with more than two responses, we consider the largest difference. As WinoGrande does not have ordered scoring, these prompts are not filtered. The number following model name indicates the fraction of the dataset used, with 1.0 indicating no filtering was performed. Experimental details about quality and diversity selection are included in Appendix~\ref{sec:appendix-hyperparmam}.

\subsection{Baselines}\label{sec:baselines}
We compare our approach that learns from heterogeneous human feedback datasets to the following fundamental baselines: Pre-trained LLM (base), Pre-trained LLM with SFT using WinoGrande, and Pre-trained LLM with SFT and RLHF using WinoGrande.
Our method uses the same heterogeneous dataset for SFT and RLHF. Eight Nvidia A100 GPUs are used for each step of the process. We test using LLaMA-7B, however our framework is naturally able to leverage any other state-of-the-art large language model. The hyperparameters and LoRA configuration are included in Appendix~\ref{sec:appendix-hyperparmam}.

\subsection{Metrics}\label{sec:metrics}
We use several metrics to measure change in gender bias, reported in Table~\ref{tab:results}. Our metrics use prompts based on the multiple choice format used in PaLM~\cite{chowdhery2022palm}, and are included in Appendix~\ref{sec:appendix-mcprompts}. Bias takes the difference in log probabilities for the correct token in WinoBias for the pro-bias and anti-bias sentences. A model reflecting no gender bias would have a difference of 0. Bias (Cluster) performs the same computation, except it considers the log probabilities for every word in the coreference cluster. This includes the pronoun used in the Bias metric, so its values are larger. Bias (Entropy) takes the relative entropy of the next token logits for the pro-bias and anti-bias sentences. This measures how different the model state is as a result of each prompt. An unbiased model would have a relative entropy of 0.

Accuracy is computed in a generative context, where the model is asked to complete a sentence. Generation is stopped after 10 new tokens or a punctuation token, whichever is sooner. Accuracy is averaged over both the pro-bias and anti-bias prompts, so this is more a measure of utility. We complement this metric with Similarity, which uses the same generation. It is how often the result, either correct or incorrect, for a pair of WinoBias sentences is shared. 

Additionally, we conduct a qualitative experiment. We ask the model to respond to several simple prompts. Evaluating models in a chatbot-like context gives another perspective on utility. The models are given single-sentence prompts such as:
\begin{quote}
    \ttfamily
     "Give me a list of top vacation destinations.\textbackslash n"
\end{quote}

\section{Results}\label{sec:results}

\subsection{Quantitative Results}\label{sec:sft-results}\label{sec:rlhf-results}
Quantitative results are reported in Table~\ref{tab:results}. Results are rounded to 4 digits after the decimal place. We find that our method is able to reduce bias by several metrics relative to all baselines, including a pre-trained model, while maintaining utility as measured by accuracy. We also see that using SFT and RLHF with our framework generally leads to lower bias than with SFT only. Based on the results for Bias (Entropy), Bias (Cluster), and Accuracy, we can get higher performance by filtering for data quality and diversity than with using the full fine-tuning dataset. We believe this result may be improved by examining more rigorous methods for measuring quality and diversity. The qualitative results also show that our framework permits the improvment on multiple measures, which are not necessarily correlated, simultaneously. With LLaMA-HD-0.6-R and LLaMA-HD-1.0-R, we achieve higher generative accuracy, a measure of utility, and higher generative similarity, a measure of bias, relative to the base model.

\subsection{Qualitative Results}

We find that the base and fine-tuned methods using WinoGrande consistently fail to follow the prompt. In many instances, the prompt is repeated indefinitely. With our method, we receive reasonable responses, likely as a result of our secondary fine-tuning dataset, OASST, including instruction-following examples. The qualitative task shows us that the method is able to train for multiple tasks at once, namely a reduction in bias and instruction following. An illustrative sample is included in Appendix~\ref{sec:appendix-qualitative}. We observe that while the base model rarely answers the prompt, LLaMA-S does on occasion respond reasonably, even though it was not explicitly instruction fine-tuned. Using only 20\% of the filtered dataset, we are able to achieve consistent instruction following (Appendix~\ref{sec:appendix-qualitative}). The highest generative accuracy and lowest bias (entropy) was also obtained by a model using a fitlered dataset, demonstrating that filtering can simultaneously improve quality and reduce bias.

\section{Conclusion} \label{sec:conc}
We find that combining datasets of heterogeneous supervision for fine-tuning can lead to performance increases beyond using only one dataset, even when the secondary dataset is less directly related to the task. We find that by varying the fraction of used data, we are able to achieve performance comparable to the full dataset, and sometimes exceed it. Most significantly, when the reduced bias result of the quantitative result are combined with the instruction-following seen in the qualitative result, we show that it is possible to fine-tune for multiple purposes simultaneously, even when the datasets include a different supervision format. Our framework can be used to improve both performance-oriented metrics, like instruction following, and to unwanted behavior like bias. This shows that it is possible to effectively fine-tune LLMs using heterogeneous supervision.

\bibliography{paper}
\appendix

\section{Appendix}\label{sec:appendix}

\subsection{Heterogeneous Dataset Creation}\label{sec:appendix-data-creation}
In this section, we include an example of heterogeneous dataset creation. The following example is from WinoGrande, our primary fine-tuning dataset. It format follows that of Section~\ref{sec:primary-finetune}, where $A_{i,0}$ is the correct response.
\begin{quote}
    \ttfamily
    $P_i$: \bsq{The box was still visible after James tried his best to manage the wrapper on it. James should have used the \_ that is small.}
    \\
    $A_{i,0}$: \bsq{box}
    \\
    $A_{i,1}$: \bsq{wrapper}
\end{quote}

The following is from OASST, our secondary fine-tuning dataset. It contains a varying number of responses to each prompt, as well as numerical labels. In the experiment, we use the score toxicity. For brevity, we only include part of the score for the first response, but each response has several numerical scores.
\begin{quote}
    \ttfamily
    $P_{i'}$: \bsq{Which affordable GPU would you recommend to train a language model?}
    \\
    $A_{i',0}$: \bsq{It heavily depends on the size...}
    \\
    $A_{i',1}$: \bsq{It is difficult to say...}
    \\
    $A_{i',2}$: ...
    \\
    $y_{i,0}$: \bsq{toxicity}=  0.00038284, \bsq{spam} = 0, ...
\end{quote}

To convert the data of type OASST into that of our primary dataset, we need to first choose two of the responses. If the prompt had only one response, it would be discarded. We select the most and least toxic responses assocaited with this prompt, and give the prompt a quality score based on that difference. For this example, we will consider this to be responses 0 and 1, with 1 the preferred response. After filtering for diversity and quality, the data entry for the unified dataset would have the following format:

\begin{quote}
    \ttfamily
    $P_{i''}$: \bsq{Which affordable GPU would you recommend to train a language model?}
    \\
    $A_{i'',0}$: \bsq{It is diffucult to say...}
    \\
    $A_{i'',1}$: \bsq{It heavily depends on the size...}
\end{quote}

Now that the format is identical to that of our primary dataset, we take the union of the primary and secondary datasets to form the homogenized dataset. The same process is applied to any additional datasets. If separate datasets were used for SFT and RLHF, the process would be repeated.

\subsection{Converting Supervision}\label{sec:appendix-converting}

In general, preference datasets can be supervised in one of three types, in order of increasing complexity: binary, ordinal, and numerical. As projecting a simpler supervision type can be noisy, we focus on simplifying supervision. In general, this is the recipe for converting a single axis of supervision:

\begin{itemize}
    \item numerical $\rightarrow$ sort increasing or decreasing, based on application $\rightarrow$ ordinal
    \item ordinal $\rightarrow$ extract best and worst responses to prompt $\rightarrow$ binary
\end{itemize}

\subsection{Hyperparameters}\label{sec:appendix-hyperparmam}
We include the hyperparameters and LoRA configuration in this section.

Quality and Diversity Selection: We use all-MiniLM-L6-v2, a sentence transformer designed to capture semantic information, to generate embeddings for each prompt~\cite{reimers-2019-sentence-bert}. The data are then separated into 10 clusters with 10 restarts using k-means. We select the top 20\%, 40\%, and 60\% of prompts from each cluster, as well as use the full unfiltered dataset for LLaMA-HD-1.0. We perform stratified random sampling to preserve the fraction of samples from each of the datasets used in the experiment, maintaining the importance of each dataset relative to the unfiltered model.

SFT: Learning Rate: 1e-5, Maximum steps: 5000, Epochs: 1, Optimizer: Adamw, LR Scheduler: cosine, Maximum text length: 512, Batch size: 4, Grad accumulation steps: 1, weight decay: 0.05. LoRA: Rank: 16, Alpha: 32, Dropout: 0.05.

Reward Model: Learning rate: 2e-5, Epochs: 1, Optimizer: Adamw, LR Scheduler: linear, Maximum text length: 512, Batch size: 4, Gradient accumulation steps: 1, weight decay: 0.001. LoRA: Rank: 8, Alpha: 32, Dropout: 0.1.

RLHF: Learning Rate: 1.41e-5, Maximum steps: 20000, Epochs: 4, Minimum generation length: 32, Maximum generation length: 128, PPO Minibatch size: 1, Batch size: 32, Gradient accumulation steps: 4. LoRA: Rank: 16, Alpha: 32, Dropout: 0.05.

\subsection{Quantitative Prompt}\label{sec:appendix-mcprompts}
We list the prompt used for our qualitative experiment here. The format is based on the multiple choice one from PaLM~\cite{chowdhery2022palm}. All quantitative metrics use this format. Details about scoring are included in Section~\ref{sec:metrics}.
\begin{quote}
    \bsq{\{sentence\} "\{pronoun\}" refers to: }
\end{quote}

\subsection{Qualitative Example}\label{sec:appendix-qualitative}

We report LLaMA-HD-0.2-S because it uses the smallest fraction of OASST, yet consistently demonstrates instruction following.

\begin{quote}
    \ttfamily
    Prompt: \bsq{What can I do in Miami, FL in November?}
    \\
    LLaMA-SFT: \bsq{I'm going to the beach in the summer...}
    LLaMA-HD-0.2-S: \bsq{ In November, you can enjoy the warm weather...}
    \\
\end{quote}

\end{document}